\documentclass[conference]{IEEEtran}
\IEEEoverridecommandlockouts

\usepackage{cite}
\usepackage{amsmath,amssymb,amsfonts}
\usepackage{algorithmic}
\usepackage{graphicx}
\usepackage{textcomp}
\usepackage{xcolor}
\usepackage{comment}
\usepackage{caption}
\usepackage{subcaption}
\usepackage{graphicx}
\usepackage{booktabs}

\def\BibTeX{{\rm B\kern-.05em{\sc i\kern-.025em b}\kern-.08em
    T\kern-.1667em\lower.7ex\hbox{E}\kern-.125emX}}
\begin{document}

\title{Clinically Inspired Symptom-Guided Depression Detection from Emotion-Aware Speech Representations \\

}

\author{\IEEEauthorblockN{Chaithra Nerella}
\IEEEauthorblockA{\textit{Speech Processing Lab, LTRC} \\
\textit{IIIT Hyderabad}\\
Hyderabad, India \\
chaithra.nerella@research.iiit.ac.in}
\and
\IEEEauthorblockN{Chiranjeevi Yarra}
\IEEEauthorblockA{\textit{Speech Processing Lab, LTRC} \\
\textit{IIIT Hyderabad}\\
Hyderabad, India \\
chiranjeevi.yarra@iiit.ac.in}

}

\maketitle

\begin{abstract}
Depression manifests through a diverse set of symptoms such as sleep disturbance, loss of interest, and concentration difficulties,etc. However, most of the exisitng works treat depression prediction either as a binary label or overall severity score without explicitly modeling symptom-specific information. This limits their ability to provide symptom-level analysis relevant to clinical screening. To address this, we propose a symptom-specific and clinically inspired framework for depression severity estimation from speech. Our approach uses a symptom guided cross attention mechanism that aligns PHQ-8 questionnaire with emotion-aware speech representations to identify which segments of a participant’s speech are more important to each symptom. To account for differences in how symptoms are expressed over time, we introduce a learnable, symptom specific parameter that adaptively controls the sharpness of attention distributions. Our results on EDAIC, a standard clinical-style dataset demonstrate improved performance outperforming prior works. Further, analyzing the attention distributions showed that higher attention was given to utterances containing cues related to multiple depressive symptoms, highlighting the interpretability of our approach. These findings outline the importance of symptom guided and emotion aware modeling for speech based depression screening.
\end{abstract}


\section{Introduction}
Depression is a growing mental health problem affecting significant popultion worldwide \cite{WHO_world2017depression}. Early detection is important as untreated cases can lead to severe loss in daily life. Clinical diagnosis often relies on standardized screening tools like Patient Health Questionnaire-8 (PHQ-8) \cite{PHQ8_kroenke2009phq}, a set of eight questions, which assesses the presence and severity of eight main symptoms. Due to the growing scale of the problem and shortage of trained mental health professionals, Automatic Depression Detection (ADD) has emerged as promising solution. ADD systems analyze behavioral information of an individual from different modalities like speech, text and video. Speech, in particular, is a natural and cost-effective compared to other modalities. Beyond the spoken words, speech carries paralinguistic cues such as prosody, pitch, etc which often change in individuals with depression, making it a valuable indicator for detection.\cite{S2_nilsonne1987acoustic} 


\begin{figure}
    \centering
    \includegraphics[width=\linewidth]
    {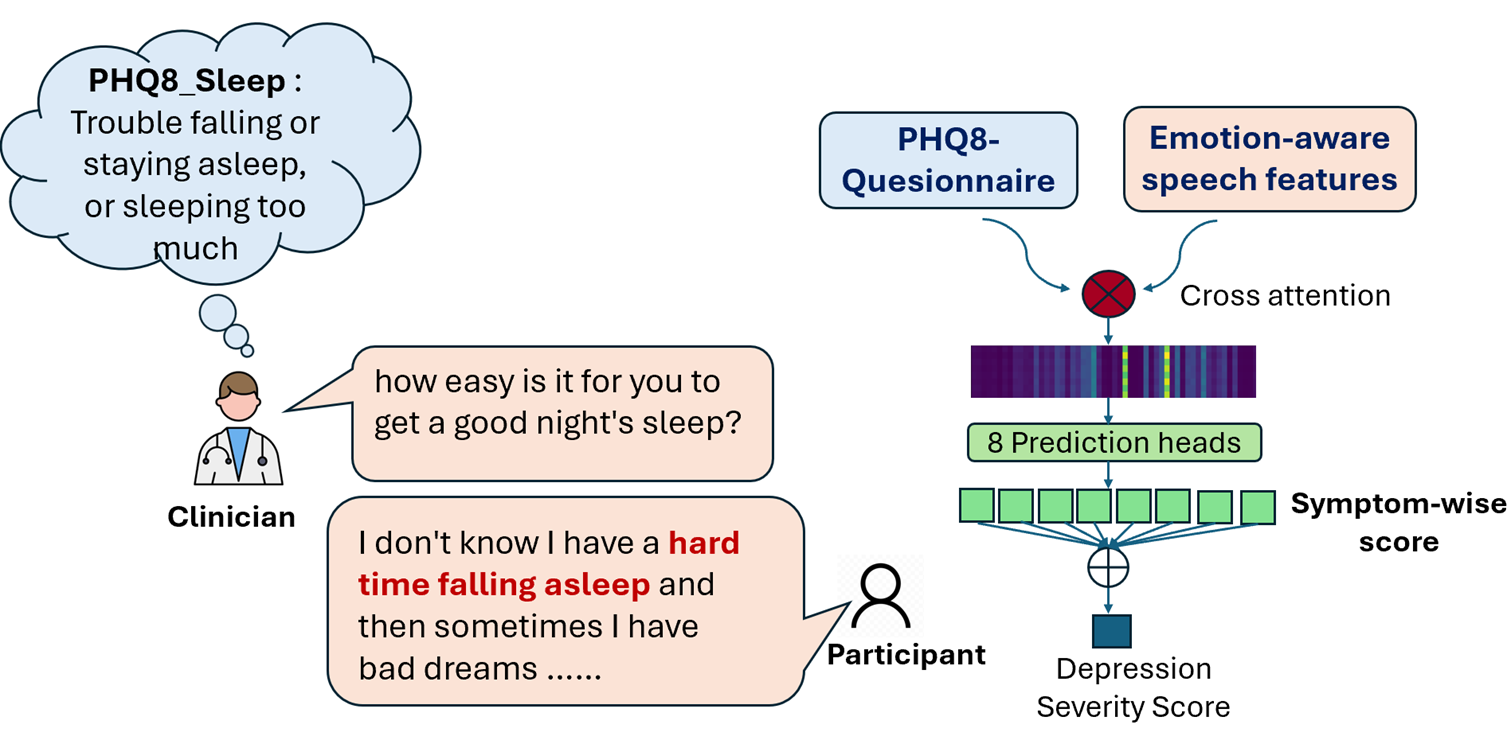}
    \caption{Proposed approach inspired from PHQ-8 based clinial practice}
    \label{fig:intro}
    \vspace{-4mm}
\end{figure}

Several studies \cite{MFCC_rejaibi2022mfcc, ssl_wu2023self} have explored acoustic features from traditional descriptors such as MFCCs  to self-supervised embeddings like wav2vec2.0, WavLM. Although they are effective, these are general-purpose representations and do not explicitly encode emotional disruptions, which are biomarkers of depressive symptoms. To overcome such limitations some works have explored multimodal approaches for depression detection by integrating speech, text, and video to capture complementary cues \cite{seneviratne2022multimodal,chen2024iifdd}.

\begin{figure*}
    \centering
    \includegraphics[width=\textwidth]{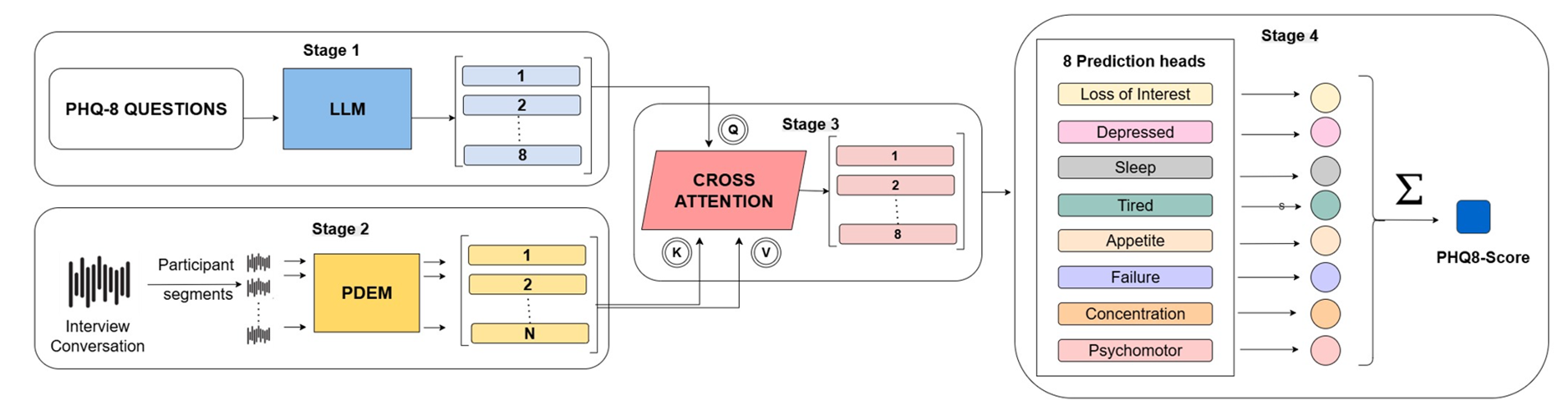}
    \caption{Block diagram of the overall framework}
    \label{fig:placeholder_1}
    \vspace{-2mm}
\end{figure*}

Despite these advancements, most of these exisiting works treat the problem as a binary classification task or as regression of a single severity score overlooking the symptom level granularity. More recent works has emphasized the importance of incorporating emotion-sensitive acoustic features for depression detection such as \cite{PDEM_wagner2023dawn} which used the Public Dimensional Emotion Model (PDEM), a Wav2Vec2-Large-Robust \cite{W2V_hsu2021robust} network trained on the MSP-Podcast \cite{MSP_lotfian2017building} dataset for emotion recognition to capture emotionally rich speech patterns relevant to depression. Other works have explored symptom level prediction by estimating PHQ-8 symptoms independently \cite{SYMP1_dumpala2024self}, \cite{SYMP2_feng2024robust}. These works demonstrate the potential of symptom-aware score prediction and its dependency on depression prediction. However, they generally treat each symptom independently and fail to explicitly model the alignment between spoken content with the PHQ-8 questionnaire, which clinicians use during their assessment as shown in Figure \ref{fig:intro}. It represents clinicians way of assessing the depression where they keep PHQ-8 symptoms in mind and ask indirectly related questions and interpret patient responses in terms of those symptoms. This kind of identifying symptom cues is crucial for for effective and interpretable depression prediction.

In order to address these gaps, in this work  we propose a clinically inspired approach that is both symptom-specific and emotion-aware. As shown in Figure \ref{fig:intro}, our approach resembles the clinicians assessment process of depression guided by PHQ-8 questionnaire. Our approach aligns PHQ-8 questionnaire with emotion-aware speech features through a symptom-guided cross-attention module. Thus, the model is guided to selectively attend to the utterances that are most relevant for each symptom. To capture the emotional patterns that are central to depression, we use PDEM model embeddings as our primary acoustic representations. Furthermore, to account for the diverse ways symptoms manifes, we introduce a learnable temperature parameter to adaptively control the sharpness of attention for each symptom. Through experiments on the EDAIC-dataset, we demonstrate that our approach outperforms prior methods achieving lower RMSE, MAE and higher CCC along with symptom-level interpretability to support clinical screening.

\vspace{-2mm}

\section{Dataset}
\textbf{EDAIC} : We carried out our experiments on EDAIC dataset which is an extended version of the DAIC-WOZ \cite{DAIC_gratch2014distress} dataset from the AVEC 2019 Detect Depression with AI Sub-challenge(DDS). 
EDAIC is a standard benchmark dataset, widely used in research on psychological distress such as depression, anxiety, and Post Traumatic Stress Disorder (PTSD) as it provides rich multimodal information for studying these conditions.
The dataset consists of 275 clinical interviews conducted with a virtual interviewer named Ellie, in a clinical-like setting. For each interview, the dataset provides audio recordings that capture vocal patterns, transcripts that capture the spoken content, and visual features from video recordings. Each participant is associated with standardized PHQ-8 scores for depression, including symptom-level scores and overall severity. Each PHQ-8 symptom is rated on a scale from 0 (not at all) to 3 (nearly every day) resulting in overall depression severity score that ranges from 0 to 24. To maintain consistency with existing works, we followed the same split which was present in the dataset: 163 participants for training, 56 for development, and 56 for testing. To the best of our knowledge, this is the only publicly available dataset that has clinical-style interviews along with PHQ-8 annotations.

\section{Methodology}

Figure \ref{fig:placeholder_1} shows the overall flow of our methodology, which has four stages. In the first stage, for every question in PHQ-8 questionnaire, we obtain semantic representations using a Large language model. Second, the spoken content from a participant is segmented into utterances using utterance level timestamps. Further, for each utterance, speech representations are obtained using PDEM model, which encodes emotion related information. Third, symptom-guided cross attention \cite{Attn_vaswani2017attention} is applied between the PDEM and PHQ8 representations to selectively focus on speech segments that are more important for the PHQ8 symptoms. Finally, all the eight question specific outputs from the  cross attention module are passed to eight prediction heads separately. Each head corresponds to a PHQ-8 symptom and outputs a score in the range [0, 3]. The individual symptom scores are then added to get the overall depression severity that ranges from [0, 24].

\subsection{Clinically Inspired Query Construction}
\label{ssec:subhead}
Clinical diagnosis of depression often relies on PHQ-8, a set of eight standardized questions that assess eight symptoms: loss of interest, feeling depressed, sleep problems, tired/fatigue, appetite, failure, concentration, and psychomotor changes. 

In our framework, we encode each question in the PHQ-8 Quessionnaire using a pretrained large language model, Roberta-large \cite{ROBERTA_liu2019roberta}. The choice of Roberta is due to its strong ability to capture contextual and semantic representations of text in natural language understanding tasks. From this encoding, we extract the CLS token embedding for each question, resulting in eight vectors, with each vector representing a symptom in a rich semantic space. These symptom-specific vectors serve as initial queries in the cross-attention module and they are treated as trainable parameters, allowing the model to adapt the semantic symptom space to the acoustic space while preserving the original clinical meaning.


    \subsection{Emotion-Aware Feature Extraction}
\label{ssec:subhead}

The spoken content from audio interview of each participant $P_i$ is segmented into utterances $P_i = [S_1, S_2, \dots, S_n]$ using the time stamps provided in the data set. Every segment $S_i$ is encoded into a 1024-dimensional embedding using the PDEM model. PDEM is a Wav2vec2-Large-Robust \cite{W2V_hsu2021robust} network fine-tuned on the MSP-Podcast\cite{MSP_lotfian2017building} dataset for emotion recognition. Unlike standard acoustic encoders, PDEM is trained to predict fine-grained emotional dimensions such as arousal, valence and dominance. Thus, by using PDEM, the extracted features also contain emotion-sensitive information, which is important as the depressed people go through a lot of emotional disturbances that could be central markers of depression.


\subsection{Symptom-Guided Cross-Attention}
\label{ssec:subhead}

In our framework, the Queries (Q) are obtained from the PHQ-8 question embeddings and Keys (K) and Values (V) come from the PDEM segment embeddings. For each participant we stack the PDEM embeddings of all speech segments into a N*1024 matrix where N represents the number of segments of that participant. The PHQ-8 question embeddings obtained from RoBERTa serve as the starting point for our queries. Rather than keeping them fixed, they are treated as trainable parameters so that the model can adapt the queries to better align with the acoustic space, while still preserving the semantics of the questionnaire. Prior to computing attention, the queries, keys, and values are normalized using LayerNorm to stabilize the training.  
The attention  between the symptom queries  and the segment embeddings is given by:
\[
\text{Attention}(Q, K, V) = \text{softmax}\!\left( \frac{QK^{\top}}{\tau \cdot \sqrt{d_k}} \right) V
\]

where $d_k$ is the embedding dimension and $\tau$ is a learnable temperature parameter associated with each symptom. $\tau$ is used to control the sharpness of attention. The smaller $\tau$ produces sharper, more selective attention (focusing on a few decisive segments), while the larger $\tau$ produces broader attention (distributing weights across many segments). This is important as different symptoms may mainfest differently. For example, 'Sleep' difficulties may be spread across multiple segments, while 'Psychomotor changes' might appear clearly in a single utterance. Thus, the model can selectively emphasize the speech segments most relevant to each PHQ-8 symptom.


\subsection{Symptom-Level Severity Estimation}
\label{ssec:subhead}
The attended representation for each symptom is passed to a small regression head consisting of two fully connected layers with ReLU and dropout. Each head outputs a single scalar score in the range [0,3], corresponding to that PHQ8-symptom. Finally, the eight predicted symptoms scores are summed to produce the overall depression severity score.
This approach resembles how clinicians assess depression, by considering each symptom individually. Predicting symptom-level scores improves the model’s interpretability and ensures it aligns with the standard clinical assessment practices.

\section{Experiments}

\subsection{Implementation details}
We trained our symptom-attention model with a batch size of 8 using the AdamW \cite{ADAMW_loshchilov2017decoupled} optimizer and a learning rate of 2e-4 . The training objective was the  Mean Squared Error (MSE) across all eight PHQ-8 symptom scores to ensure that the model simultaneously learns the symptom-level predictions. Each prediction head has two fully connected layers with hidden dimensions of 128 and 1 with a dropout of 0.1 in between to prevent overfitting. We used gradient clipping with a maximum norm of 1.0 to stabilize training and prevent exploding gardients. The model was trained for 20 epochs and the checkpoint corresponding to the lowest RMSE on the development set for the total PHQ-8 score  was selected for evaluation on the test set.

\subsection{Evaluation metrics}
We evaluated performance at two levels by predicting individual symptom scores and the overall severity. Following the works of \cite{fan2019multi} \cite{yu2025using}, we used three metrics to evaluate: Root Mean Squared Error (RMSE), Mean Absolute Error (MAE), and Concordance Correlation Coefficient (CCC). RMSE captures larger errors, MAE reflects the average prediction error, and CCC measures both correlation and agreement with the ground truth. These metrics help us understand the reliability of the model's predictions which is necessary for clinical applications.

\textbf{Baselines:} We compare our approach with the existing works in Table \ref{tab:comparison}. In particular, we considered the approach of \cite{yu2025using} as our primary baseline as it also uses PDEM model for extracting the speech representations but it depends on a 
distance based selection of emotionally rich segments using arousal, valence and dominance scores given by the PDEM model. Though it helps to identify emotionally expressive segments, it does not explicitly consider whether the segments are relevant for depressive symptoms. Thus its predictions may capture emotional intensity without clinically relevant symptom cues.

However, in our method, we use symptom-specifc cross attention mechanism to directly align PHQ8 quesionnaire embeddings with the speech representation. This helps to focus on both emotionally rich and symptom specific utterances.

\section{Results and Discussion}

\subsection{Quantitative Evaluation}

\begin{table}[htbp]
\caption{Comparison with prior works on EDAIC}
\begin{center}
\label{tab:comparison}
\begin{tabular}{lccc}
\hline
\textbf {Method} & \textbf{RMSE} & \textbf{MAE} & \textbf{CCC} \\
\hline
Zhang \textit{et al.} \cite{zhang2019evaluating} & 6.78 & 5.77  & --  \\
Fan \textit{et al.} \cite{fan2019multi} & 5.91 & 4.39  & 0.43 \\
Yin \textit{et al.} \cite{yin2019multi} & 5.50 & --  & 0.44 \\
Zhao \textit{et al.} \cite{zhao2022unaligned} & 5.78 & 5.13 & -- \\
Makiuchi  \textit{et al.}  \cite{rodrigues2019multimodal} & 6.11 & -- & 0.40 \\
Han \textit{et al.} \cite{spatial} & 6.29 & 5.38  & --\\
Yu \textit{et al.} Baseline\cite{yu2025using}  & 5.58 & -- & 0.48 \\
\textbf{Ours} & \textbf{5.15}  & \textbf{4.13} & \textbf{0.52} \\
\hline
\end{tabular}
\end{center}
\end{table}

\begin{table}[!htbp]
\centering
\caption{Sypmtom-wise Performance Metrics}
\label{tab:symptom_metrics}
\begin{tabular}{lrr}
\toprule
\hline
\textbf{Symptom} & \textbf{RMSE} & \textbf{MAE} \\
\hline
\midrule
No Interest      & 0.891         & 0.638        \\
Depressed        & 0.728         & 0.571        \\
Sleep            & 1.104         & 0.903        \\
Tired            & 1.005         & 0.842        \\
Appetite         & 1.008         & 0.800        \\
Failure          & 0.903         & 0.734        \\
Concentration    & 0.893         & 0.708        \\
Psychomotor      & 0.738         & 0.592        \\
\hline
\bottomrule
\end{tabular}
\end{table}

\textbf{Comparison with prior works:} Table \ref{tab:comparison} compares the proposed method against the state of the art methods on EDAIC test set. The results show that our method achieves the best results across all three metrics outperforming existing works. 
To highlight these improvements, we compare with prior methods by Zhang et al \cite{zhang2019evaluating} that used general acoustic and linguistic cues and  Yu et al. \cite{yu2025using} that relied only on emotionally rich segments. The higher CCC of 0.52, corresponding to 8.3\% relative improvement over baseline further indicates that modeling depression at the clinical symptom level gives more reliable severity estimation. 

Two main factors contributed to this improvement. First, modeling at symptom level by explicitly aligning with PHQ-8 Quessionnaire allows the model to focus on utterances that are relevant to each clinical symptom rather than only on emotionally rich segments. Second, emotion-aware speech representations capture cues that are closely linked to depressive symptoms. This indicates that combining symptom-guided attention with emotion-aware speech representations is useful for the clinically aligned depression prediction.

\textbf{Symptom-wise results:} We further report symptom-wise performance in Table \ref{tab:symptom_metrics} to show variation in performace across different symptoms. The model achieved the lowest error rates for the symptoms Depressed mood (RMSE = 0.280, MAE = 0.571) and Psychomotor changes (RMSE = 0.375, MAE = 0.592), indicating that these symptoms can be detected reliably through speech.

In contrast, the model gave high errors for the physiological symptoms, particularly symptoms related to Sleep disturbances (RMSE=1.104, MAE=0.903) and Tiredness (RMSE=1.005, MAE=0.842). The results are consistent with clinical intuition as symptoms related to internal physiological states are not always expressed explicitly in short speech segments. Even though if individuals report verbally about problems with sleep or energy, this information is very difficult to detect by speech-only systems. These kind of symptoms require complementary modalities like facial cues to predict effectively.

\begin{table}[!htbp]
\centering
\caption{Performance comparison with different $\tau$ settings}
\label{tab:my-table}
\resizebox{\columnwidth}{!}{%
\begin{tabular}{|l|l|l|l|}
\hline
\textbf{$\tau$ setting}                         & \textbf{RMSE} & \textbf{MAE}  & \textbf{CCC}  \\ \hline
{No $\tau$}                      & {5.36} & {4.22} & \textbf{0.55} \\ \hline
{Learnable $\tau$(global)}      & {5.31} & {4.17} & {0.54} \\ \hline
{Learnable $\tau$( per symptom)} & \textbf{5.15} & \textbf{4.13} & {0.52} \\ \hline
\end{tabular}%
}
\vspace{-1mm}
\end{table}

\textbf{Effect of Symptom-specific attention scaling $\tau$ setting:} Table \ref{tab:my-table} presents the performance of our approach under different configurations of the temperature parameter $\tau$ in the attention mechanism. In the absence of temperature scaling, the model yields an RMSE of 5.36, MAE of 4.22 and CCC of 0.55. Introducing a single learnable global $\tau$ slightly improves the results, reducing RMSE to 5.31 , MAE to 4.22 and 0.54. The best performance is observed when $\tau$ is learned independently for each symptom, achieving an RMSE of 5.15 and an MAE of 4.13.

The per-symptom learnable $\tau$ allows the model to adjust its attention for each symptom, reflecting that some symptoms may be expressed across multiple speech segments while others may appear in few segments. The CCC values remain high across all settings when compared with existing works indicating that the model predictions are aligned with the groud truth depression severity.

\subsection{Interpretability}
\label{ssec:subhead}

\begin{figure}[t]
    \centering
    \begin{subfigure}{\linewidth}
        \centering
        \includegraphics[width=\linewidth]{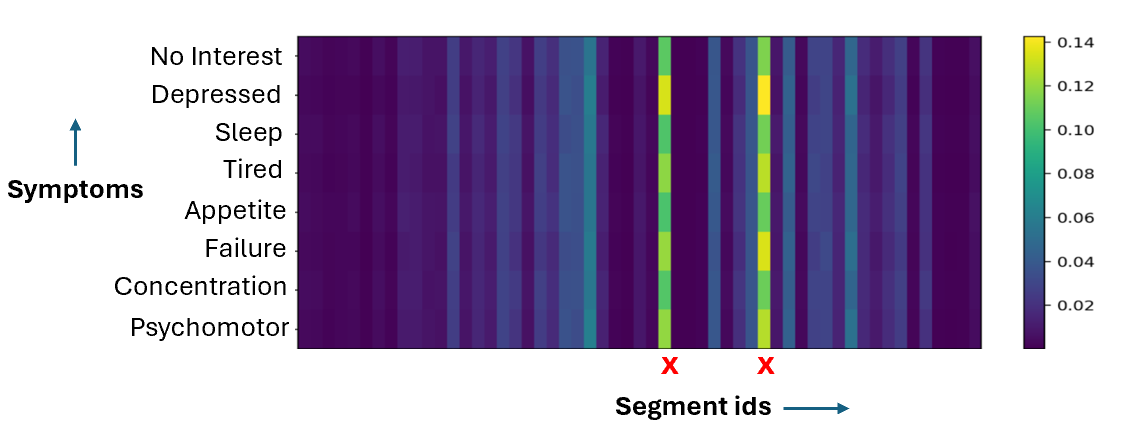}
        \caption{}
        \label{fig:attn}
    \end{subfigure}

    \begin{subfigure}{\linewidth}
        \centering
        \includegraphics[width=\linewidth]{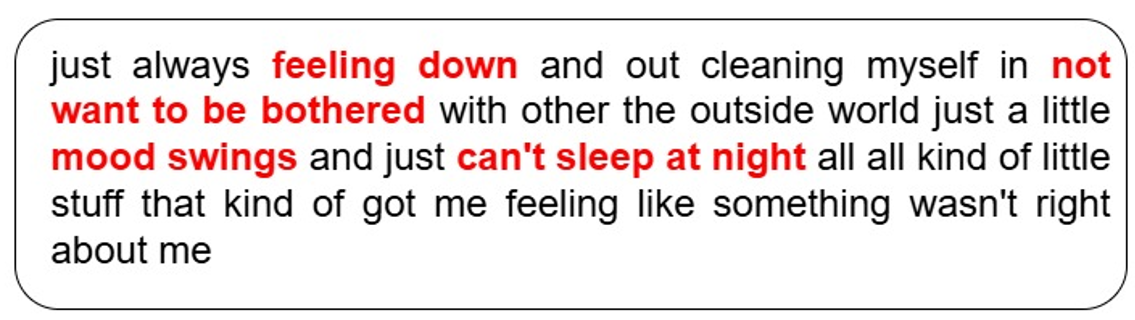}
        \caption{}
        \label{fig:placeholder_2}
    \end{subfigure}
    
    
    \begin{subfigure}{\linewidth}
        \centering
        \includegraphics[width=\linewidth]{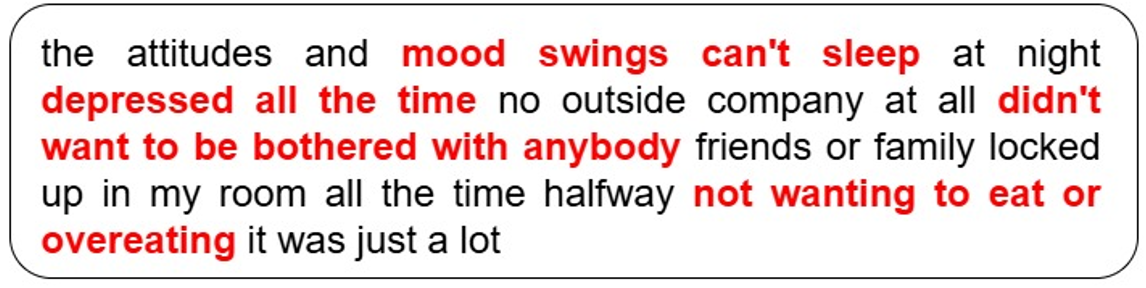}
        \caption{}
        \label{fig:placeholder_3}
    \end{subfigure}
    \vspace{-3mm}
    \caption{Visualization of attention heatmap and text alignment of the sentences marked with 'x'\\  }
    \label{fig:attn_text}
    \vspace{-8mm}
\end{figure}

For clinically related tasks, it is important that the models are interpretable in addition to high accuracy or score. To understand how our model is emphasizing on symptom relevant information, we analysed the attention weights from the cross attention module. Figure \ref{fig:attn} shows the attention heatmap for a participant, showing how the model distributes its focus across different speech segments. Although attention is spread across several utterances, two segments (marked with 'x' ) receive higher attention weights comparatively, indicating that these segments are most important for predicting depression.

Figure  \ref{fig:placeholder_2} and \ref{fig:placeholder_3} shows the text content of these two utterances.
In Figure \ref{fig:placeholder_2}, the high attention is obtained from the following : “can’t sleep at night,” “feeling depressed all the time,” “not wanting to be bothered,” and “not wanting to eat or overeating.” These expressions are directly related to four symptoms of PHQ-8: sleep disturbance, depressed mood, loss of interest, and appetite changes. Similarly, in Figure \ref{fig:placeholder_3}, the high attention is obtained from "feeling down," "not want to be bothered with anybody," "mood swings," "can't sleep at night" which corresponds to multiple PHQ-8 symptoms. In both the segments these are cues of multiple PHQ-8 symptoms, thus they receive stronger attention from the model. This symptom-selective attention mechanism shows that the proposed cross-attention module effectively learns to identify and focus on clinically meaningful utterances that correspond to PHQ-8 symptoms. This level of interpretability makes the framework more trustworthy to support clinicians to use for screening.

Further examination reveals that all symptoms do not attend equally to the same utterance; some assign stronger focus while others remain less responsive. In the marked sentences for this participant, this difference is particularly evident, and similar patterns are observed in other participants as well. This variation highlights the role of the learnable temperature parameter ($\tau$), which controls the sharpness of the attention and facilitates the specific focus of symptoms.


\section{Conclusion}


 In this work, we presented a symptom-specific and clinically inspired attention framework for automatic depression detection that aligns participant speech directly with the eight PHQ-8 symptoms. By incorporating emotion-sensitive information into the acoustic features, the approach captures subtle emotional patterns that are closely associated with depression. The symptom-guided attention enables the model to focus more on utterances that are relevant to depressive symptoms. Experiments on the E-DAIC dataset showed that this approach achieves strong performance along with interpretability. For future work, we plan to integrate  other modalities such as text transcripts and video recordings along with speech to assess all the symptoms more effectively.

\bibliographystyle{IEEEtran}


\end{document}